\documentclass{article}
\usepackage{ijcai23}

\usepackage{multirow}
\usepackage{times,bm}
\usepackage{soul}
\usepackage{url}
\usepackage{subfigure}
\usepackage{graphicx}
\usepackage{amsmath}
\usepackage{booktabs}
\usepackage{algorithm}
\usepackage{algorithmic}
\usepackage{amssymb,amsfonts}
\usepackage{amsthm,amsmath,amssymb}

\usepackage{color}
\usepackage{helvet}
\usepackage{courier}
\usepackage{bbm}
\usepackage{makecell}

\usepackage[hidelinks]{hyperref}
\usepackage[utf8]{inputenc}
\usepackage[small]{caption}
\usepackage[switch]{lineno}
\usepackage{amsfonts,amssymb}
\usepackage{enumitem}

\title{Adaptive Path-Memory Network for Temporal Knowledge Graph Reasoning}

\author{
Hao Dong$^{1,2}$\footnotemark[1]
\and
Zhiyuan Ning$^{1,2}$\footnotemark[1]
\and
Pengyang Wang$^3$\and
Ziyue Qiao$^4$\and
Pengfei Wang$^{1,2}$\footnotemark[2] \and\\
Yuanchun Zhou$^{1,2}$\And
Yanjie Fu$^5$ \\
\affiliations
$^1$Computer Network Information Center, Chinese Academy of Sciences, Beijing\\
$^2$University of Chinese Academy of Sciences, Beijing\\
$^3$State Key Laboratory of Internet of Things for Smart City, University of Macau, Macau\\
$^4$The Hong Kong University of Science and Technology (Guangzhou), Guangzhou\\
$^5$Department of Computer Science, University of Central Florida, Orlando\\
\emails
donghcn@gmail.com,
ningzhiyuan@cnic.cn,
pywang@um.edu.mo,
ziyuejoe@gmail.com,\\
\{pfwang, zyc\}@cnic.cn,
yanjie.fu@ucf.edu
}

\begin{document}

\maketitle

\renewcommand{\thefootnote}{\fnsymbol{footnote}}
\footnotetext[1]{Equal contribution.}
\footnotetext[2]{Corresponding author.}
\renewcommand{\thefootnote}{\arabic{footnote}}

\begin{abstract}

Temporal knowledge graph (TKG) reasoning aims to predict the future missing facts based on historical information and has gained increasing research interest recently. 
Lots of works have been made to model the historical structural and temporal characteristics for the reasoning task.
Most existing works model the graph structure mainly depending on entity representation. 
However, the magnitude of TKG entities in real-world scenarios is considerable,
and an increasing number of new entities will arise as time goes on.
Therefore, we propose a novel architecture modeling with relation feature of TKG, namely 
a\textbf{DA}ptiv\textbf{E} path-\textbf{M}em\textbf{O}ry \textbf{N}etwork (DaeMon), 
which adaptively models the temporal path information between query subject and each object candidate across history time. It models the historical information without depending on entity representation. 
Specifically, DaeMon uses path memory to record the temporal path information derived from path aggregation unit across timeline considering the memory passing strategy between adjacent timestamps. 
Extensive experiments conducted on four real-world TKG datasets demonstrate that our proposed model obtains substantial performance improvement and outperforms the state-of-the-art up to 4.8\% absolute in MRR.  

\end{abstract}

\section{Introduction}
Knowledge graphs (KGs) are multi-relational graphs that represent real-world facts and events. They are composed of nodes that represent entities and edges that represent the relationships between those entities. 
The edges are typically structured as triples in the form of (subject, relation, object), such as \textit{(Obama, visit, China)}. 
Temporal Knowledge Graphs (TKGs) are a special class of KGs that account for the temporal evolution of knowledge. 
Specifically, TKGs represent each fact as a quadruple that includes a timestamp, i.e., (subject, relation, object, timestamp), offering novel insights and perspectives on multi-relational data that varies over time. 
Recently, TKGs have been widely applied in a range of contexts, like policy-making, stock prediction, and dialogue systems. 
The task of TKG reasoning involves the inference of new facts from known ones, which can be performed under two primary settings: interpolation and extrapolation~\cite{jin2020Renet}. 
The interpolation setting seeks to fill in missing facts within a specified range of timestamps, while extrapolation attempts to predict future events based on historical knowledge~\cite{jin2019recurrent,trivedi2017know}. 
This study focuses on the latter setting, specifically designing a model for the prediction of links at future timestamps.

\begin{figure}
    \centering
    \includegraphics[width=.40\textwidth]{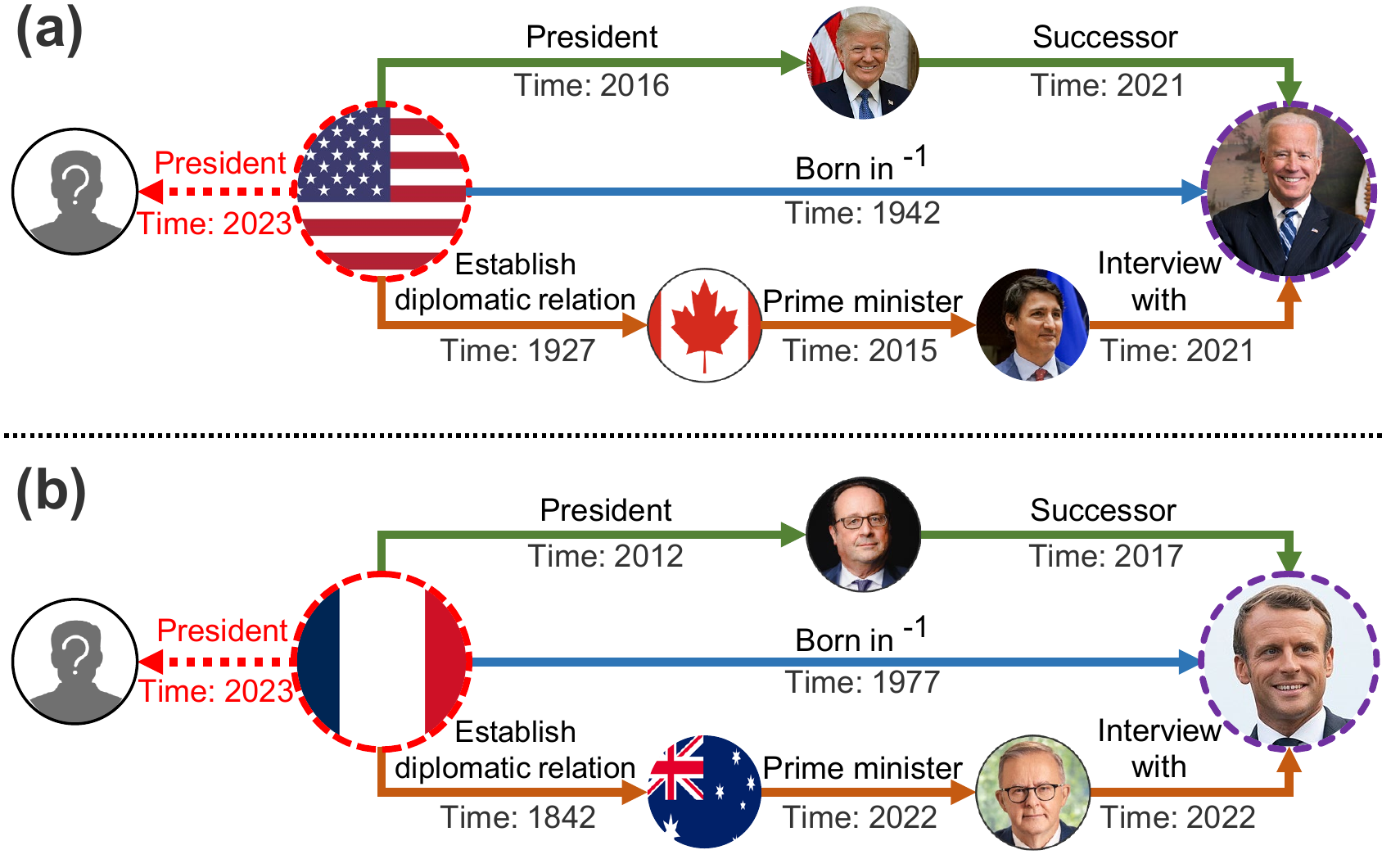}
    \caption{An Example of Temporal Path. (a) and (b) have different entities appearing in history, but the same temporal paths between query subject (\textit{United States} / \textit{France}) and an object candidate (\textit{Joe Biden} / \textit{Emmanuel Macron}).}
    \label{fig:intro}
    \vspace{-4mm}
\end{figure}

In order to make full use of historical fact information~\cite{sanchez2020learning}, various attempts have been made for TKG reasoning task to model the structural and temporal characteristics of TKGs.
In these approaches, \textit{\textbf{entities}} that appear throughout history are crucial components for modeling both structural and temporal characteristics:
(a) For structural characteristics, the identification of the entities and their respective neighbor entities are essential, as most current models utilize structural information extracted from historical facts as the basis for TKG reasoning. 
Modeling graph structures requires the use of entities (i.e., the nodes).
For example, RE-NET~\cite{jin2020Renet} employs RGCN-based aggregator for message passing between adjacent entities, while CyGNet~\cite{zhu2021learning} introduces a copy-generation mechanism to track the distribution of historical neighbors for the query entity.
(b) For temporal characteristics, entities serve as a crucial link in connecting subgraphs from different periods. 
Examples of this include a series of RNN-based methods that model the dynamic evolution of entities over time to describe temporal relationships~\cite{li2021temporal}, as well as approaches such as xERTE~\cite{han2020explainable} and TITer~\cite{sun2021timetraveler} that use entities and their neighbor entities as bridges to weave individual subgraphs into an interlinked TKG therefore considering the temporal information. 
However, the magnitude of TKG entities in real-world scenarios is considerable. 
If all historical entities are incorporated into the model, the associated overhead would be substantial. 
In addition, as TKGs continue to evolve, an increasing number of new entities will arise over time. 
Reasoning on new entities that have not been trained in the absence of historical context also presents a complex problem.

In this study, we will concentrate on the examination of \textit{\textbf{temporal path}} as a means of utilizing historical event information in a novel and \textit{\textbf{entity-independent}} manner.
There exist numerous temporal paths following the timeline in TKGs, such as \textit{Establish\_Diplomatic\_Relation} → \textit{Prime\_Minister} → \textit{Interview\_With} in Figure~\ref{fig:intro}. 
By considering the whole temporal paths which exist between the query entity and any candidate entities, we are able to infer the potential relationships that may transpire between them in the future.  
For instance, Figure~\ref{fig:intro}(a) shows a query \textit{(United States, President, ?, 2023)} and the corresponding set of temporal paths between the query entity \textit{United\_States} and a candidate entity \textit{Joe Biden}, while Figure~\ref{fig:intro}(b) shows another query \textit{(France, President, ?, 2023)} and the corresponding set of temporal paths between the query entity \textit{France} and a candidate entity \textit{Emmanuel Macron}. 
Although the entities appearing in the history event information corresponding to these two different queries are not identical, the pattern of the same temporal paths between the query entity and the candidate entity will deduce the same conclusion. 
That is, the same set of temporal paths \{\textit{Born\_In$^{-1}$}, \textit{President} → \textit{Successor}, \textit{Establish\_Diplomatic\_Relation} → \textit{Prime\_Minister} → \textit{Interview\_With}\} between \textit{United States}/\textit{France} and \textit{Joe Biden}/\textit{Emmanuel Macron} will help get the same conclusion \textit{(United States/France, President, Joe Biden/Emmanuel Macron, 2023)}.
It is noteworthy that the identity of the intermediate entities connected by temporal paths and whether they are new entities or pre-existing entities, has no impact on the ultimate prediction, thus rendering such a modeling approach completely entity-independent. 

Therefore, we try to model all temporal paths between the query subject entity and candidate object entities and propose a novel architecture modeling with relation features over timeline. 
Specifically, it includes 3 components, namely, 
(1) \textbf{Path Memory}: to record the comprehensive paired representations/embeddings of all {temporal paths} between the query entity and any candidate entities. 
(2) \textbf{Path Aggregation Unit}: to capture paired representations of {paths} from a subgraph of single timestamp. 
(3) \textbf{Memory Passing Strategy}: as time progresses, the representation of {temporal paths} in a subgraph from single timestamp is captured in a sequential manner, and is then utilized to update the stored representation of {temporal paths} in memory. 
Our main contributions are as follows:
\begin{itemize}[leftmargin=*]
\item\noindent We propose an adaptive path-memory network for temporal extrapolated reasoning over TKGs, by collecting the path information related to specific queries, considering the structural information of subgraph and temporal characteristics among subgraph sequences.
\item\noindent In order to avoid the limitations of entity-dependent model, we develop DaeMon based on the relation feature to adaptively capture the path information between entities without considering the entity representation, which is more flexible and efficient.
\item\noindent Extensive experiments indicate that our proposed model substantially outperforms existing state-of-the-art results and achieves better performance (up to 4.8\% absolute improvement in MRR) over commonly used benchmarks, meanwhile with substantial migration ability.
\end{itemize}

\begin{figure*}[!ht]
    \centering
    \includegraphics[width=1\textwidth]{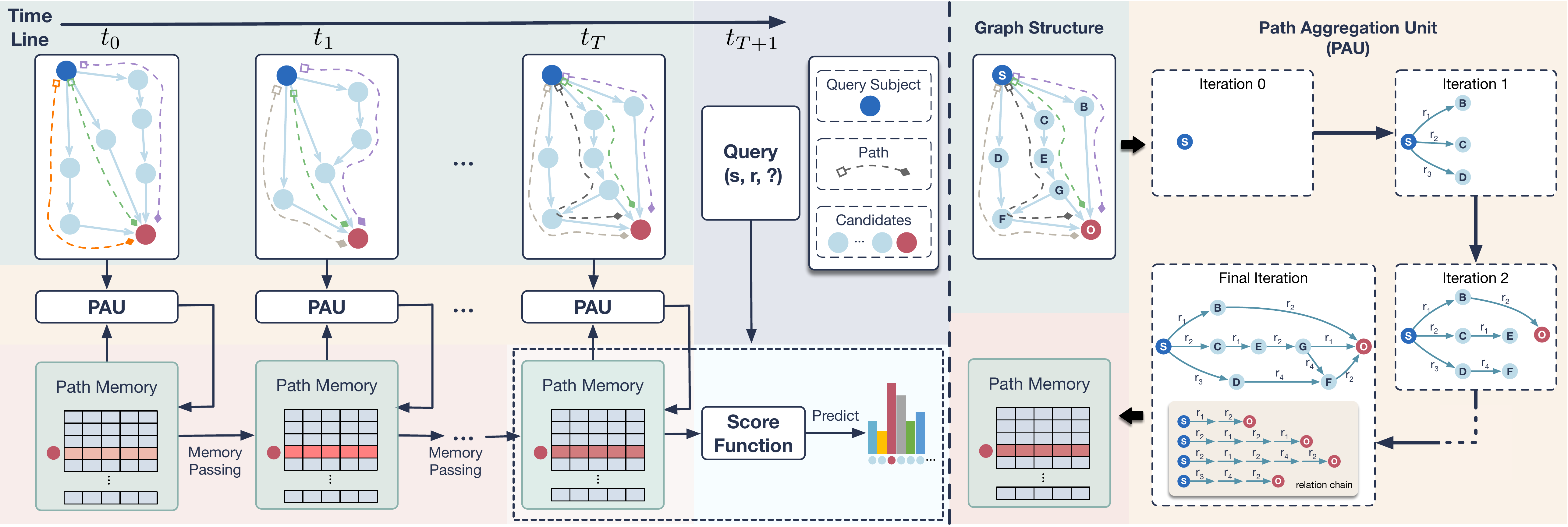}
    \caption{Framework Overview. We take the modeling temporal path between query subject (dark blue node) and a selected object candidate (dark red node) as an example, and show path memory update process only focusing on the selected object candidate for the sake of clarity. }
    \label{fig:framework}
    \vspace{-4mm}
\end{figure*}

\section{Related Work}
\subsection{Static KG Reasoning}
The development of knowledge graph embedding (KGE) methods, which aim to embed entities and relations into continuous vector spaces~\cite{bordes2013translating,2019RotatE,li-etal-2022-transher}, has garnered significant interest in recent years. 
These methods can be classified into various categories, including translational models, semantic matching models, and deep neural network approaches utilizing feed-forward or convolutional layers. 
Recent research has focused on the development of GNN-based models such as GCN~\cite{welling2016semi}, R-GCN~\cite{schlichtkrull2017modeling}, WGCN~\cite{shang2019end}, VR-GCN~\cite{ye2019vectorized}, and CompGCN~\cite{vashishth2019composition}, which combine content and structural features in a graph and jointly embed nodes and relations in a relational graph~\cite{Ning2021LightCAKEAL,qiao2020context}. 
Additionally, path-based methods have been widely adopted for KG reasoning~\cite{zhu2021neural}.
While these approaches have proven effective in static KGs ignoring the time information, they are unable to predict future events. 

\subsection{Temporal KG Reasoning}
The development of temporal KG reasoning models has garnered significant attention in recent years, particularly in regard to both interpolation and extrapolation scenarios. 
Interpolation models such as ~\cite{dasgupta2018hyte} and ~\cite{wu2020temp} aim to infer missing relations within observed data, but are not equipped to predict future events beyond the designated time interval. 
In contrast, extrapolation models have been proposed to predict future events, such as Know-Evolve~\cite{trivedi2017know}, which utilizes temporal point process to model continuous time domain facts but cannot capture long-term dependencies. 
The copy-generation mechanism employed by CyGNet~\cite{zhu2021learning} allows for the identification of high-frequency repetitive events.
xERTE~\cite{han2020explainable} offers understandable methods for their predictions yet have a limited scope of application. 
TANGO~\cite{han2021learning} uses neural ordinary differential equations to represent and model TKGs.
Other models, such as RE-NET~\cite{jin2020Renet} and RE-GCN~\cite{li2021temporal}, utilize GNN or RNN architectures to capture temporal patterns. 
While progress has been made in the field, there remains a need for TKG reasoning models that are both flexible and efficient, able to handle long-term dependencies and scalable demand. 


\section{Problem Formulation}
A temporal knowledge graph $\mathcal{G}$ can be viewed as a multi-relational, directed graph with time-stamped edges between entities.
We can formalize a TKG as a sequence of knowledge graph snapshots ordered by timestamp, i.e., $\mathcal{G}=\{G_1,G_2,...,G_t,...\}$.
A fact in a subgraph $G_t$ can be represented as a quadruple $(s,r,o,t)$ or $(s_t,r_t,o_t)$.
It describes that a fact of relation type $r_t \in \mathcal{R}$ occurs between subject entity $s_t \in \mathcal{E}$ and object entity $o_t \in \mathcal{E}$
at timestamp $t \in \mathcal{T}$, where $\mathcal{E}$, $\mathcal{R}$ and $\mathcal{T}$ denote the finite sets of entities, relations, and timestamps, respectively. We need to state in advance is that we will use \textbf{bold} items to denote vector representations in later context.

The extrapolation reasoning task aims to predict the missing object entity $o$ via answering query like $(s,r,?,t_q)$ with historical
known facts $\{(s,r,o,t_i)|t_i<t_q\}$ given, while the facts in the query time period $t_q$ is unknown. 
Specifically, we consider all entities in $\mathcal{E}$ as the candidates and 
rank them by the score function to predict missing object of the given query.
Besides, for each fact in TKG $\mathcal{G}$, we add inverse quadruple $(o,r^{-1},s,t)$ into $G_t$, correspondingly.
Note that, when predicting the missing subject of a query $(?,r,o,t_q)$, we can convert it into $(o,r^{-1},?,t_q)$.
Without loss of generality, we describe our model as predicting the missing object entity.

\section{Methodology}
In this section, we introduce the proposed model, Adaptive Path-Memory Network (DaeMon), 
as illustrated in Figure \ref{fig:framework}. We start with an overview and explain model architecture as well as its training and inference procedures in detail. 

\subsection{Model Overview}
The main idea of DaeMon is to adaptively model the query-aware path information between query subject $s$ and each object candidate $o_i \in \mathcal{E}$ across history time with sequential subgraph while considering the query relation $r$.
Since subgraphs of each timestamp are independent and unconnected, we hope to capture the connection information across time by updating the collected path information constantly.
To record the path information across timestamps, we construct \textbf{path memory} to keep the status of current collected information temporarily. And then we use \textbf{path aggregation unit} to process and model comprehensive path information between $s$ and $o_i \in \mathcal{E}$ based on the memory status. Furthermore, we propose \textbf{memory passing strategy} to guide the memory status passing between adjacent times. Based on the final learned query-aware path information stored in memory, reasoning at the future timestamps can be made with a matching function.
A high-level idea of modeling path information is to capture the relation chain pattern among the subgraph sequence.

\subsection{Path Memory}\label{method_memory}
To keep the connection information between query subject $s$ and each object candidate $o_i \in \mathcal{E}$ along the timeline, we construct query-aware memory unit to cache the obtained path information in the past and use it for subsequent processing, which can make the later processing refine the previous information continuously, considering the specific query $(s,r,?)$. We denote the status of memory as $\textbf{M}_{(s,r)\rightarrow \mathcal{E}}^{t} \in \mathbb{R}^{|\mathcal{E}| \times d}$ at timestamp $t$, where $|\mathcal{E}|$ is the cardinality of candidates set and $d$ is the dimension of path embedding.

Specifically, given a query $(s,r,?)$, we take an object candidate $o_i$ as an example. The path information between $s$ and $o_i$ we consider is the aggregation form of all paths that start with $s$ and end with $o_i$. Since our goal is to capture the path information across history timeline, we can finally obtain complete path information representation $\textbf{m}_{(s,r)\rightarrow o_i}^{T} \in \textbf{M}_{(s,r)\rightarrow \mathcal{E}}^{T}$ stored in memory at the last history timestamp $T$. 

Formally, we describe $\textbf{m}_{(s,r)\rightarrow o_i}^{t}$ at timestamp $t$ as follow:
\begin{equation}\label{eq_overview}
\textbf{m}_{(s,r)\rightarrow o_i}^{t} = 
{MPS}(
\textbf{m}_{(s,r)\rightarrow o_i}^{t-1})
\oplus
{PAU}(\mathcal{P}_{s \rightarrow o_i}^t),
\end{equation}
\begin{equation}\label{eq_agg_intro}
{PAU}(\mathcal{P}_{s \rightarrow o_i}^t) =
\textbf{p}_0^t \oplus \textbf{p}_1^t \oplus \cdot\cdot\cdot \oplus \textbf{p}_{|\mathcal{P}_{s \rightarrow o_i}^t|}^t | _{{\rm p}_k^t\in \mathcal{P}_{s \rightarrow o_i}^t},
\end{equation}

\noindent where $MPS(\cdot)$ denotes the memory passing strategy, which is designed to inherit memory status of the previous timestamp and $PAU(\cdot)$ is path aggregation unit that aggregates paths information between query subject $s$ and object candidate $o_i$, which are introduced in the following section respectively; $\mathcal{P}_{s \rightarrow o_i}^t$ denotes the set of paths from $s$ to $o_i$ at timestamp $t$ and $\oplus$ denotes the paths aggregation operator. A path $\textbf{p}_k^t \in \mathcal{P}_{s \rightarrow o_i}^t$ is defined as follow when it contains relation chain edges as $(e_0 \rightarrow e_1 \rightarrow \cdot\cdot\cdot \rightarrow e_{|\textbf{p}_k|}^t)$:
\begin{equation}\label{eq_msg_intro}
\textbf{p}_k^t = \textbf{w}_r(e_0) \otimes \textbf{w}_r(e_1) \otimes \cdot\cdot\cdot \otimes \textbf{w}_r(e_{|{\rm p}_k^t|}),
\end{equation}
\noindent where $e_{(1,2,...,|{\rm p}_k^t|)}$ denotes the edge types in path ${p}_k^t$, $\textbf{w}_r(e_*)$ is the $r$-aware representation of edge $e_*$ and $\otimes$ denotes the operator of merging edges information in the path ${\rm p}_k^t$.

\subsection{Path Aggregation Unit}\label{method_pau}

Path aggregation unit (PAU) is to process the representation of path information between a specific subject $s$ and all candidates $o_i \in \mathcal{E}$, while considering the specific query relation type $r$. That means we only focus on features related to the query $(s,r,?)$. 

We design a temporal path aggregation approach to constantly learn the history subgraphs structure over timeline. Since path feature in each subgraph is static without considering temporal feature, PAU learns path feature locally at each timestamp based on the memory status of previous adjacent timestamp and adaptively refines the memory with the development of time. 
Formally, given a query $(s,r,?)$, path information of $\mathcal{P}_{s \rightarrow \mathcal{E}}^t$ for all $o_i \in \mathcal{E}$ will be learned by $w$-layers $r$-aware path aggregation neural network, based on previous memory status and topology of the subgraph at timestamp $t$. Memory status $\textbf{M}_{(s,r)\rightarrow \mathcal{E}}^{t}$ will be updated as aggregated path information, after finishing $w$ layers aggregation iteration.

We can use $\textbf{H}_{(s,r)\rightarrow \mathcal{E}}^{t,l} \in \mathbb{R}^{|\mathcal{E}| \times d}$ to denote the $l$-th layer iterative path information status, and $\textbf{h}_{(s,r)\rightarrow o_i}^{t,l} \in \textbf{H}_{(s,r)\rightarrow \mathcal{E}}^{t,l}$ denotes the status of candidate $o_i$ at the $l$-th iteration, where $l \in [0, w]$ and $d$ is the dimension of path embedding. When $l=0$, $\textbf{H}_{(s,r)\rightarrow \mathcal{E}}^{t,0}$ denotes the initial status of iterations at timestamp $t$.
For simplicity, we abbreviate $\textbf{H}_{(s,r)\rightarrow \mathcal{E}}^{t,l}$, $\textbf{h}_{(s,r)\rightarrow o_i}^{t,l}$ 
as $\textbf{H}_{\mathcal{E}}^{t,l}$, $\textbf{h}_{o_i}^{t,l}$, respectively, 
and it is still representing path information for a specific query subject $s$ and relation $r$ rather than entity representation. 
Here we take $\textbf{h}_{o_i}^{t,l}$ as an example to describe the PAU iterative process at timestamp $t$ as follow:
\begin{equation}
\begin{split}
\textbf{h}_{o_i}^{t,l} = \textsc{Agg} \bigg(
\Big\{
\textsc{Msg}\big(\textbf{h}_{z}^{t,l-1},\textbf{w}_r(z,p,o_i)\big) \\
\Big|(z,p,o_i)\in G_t 
\Big\}
\bigg),
\end{split}
\end{equation}

\noindent where $\textsc{Agg}(\cdot)$ and $\textsc{Msg}(\cdot)$ are the aggregation and message generation function, corresponding to the operator in Equation \ref{eq_agg_intro} and \ref{eq_msg_intro}, respectively; 
$\textbf{w}_r$ denotes $r$-aware representation of edge type $p$; $(z,p,o_i)\in G_t $ is an edge in subgraph $G_t$ that relation $p$ occurs between an entity $z$ and candidate entity $o_i$. 

Moreover, we initialize all relation type representation with a learnable parameter $\textbf{R} \in \mathbb{R}^{|\mathcal{R}| \times d} $ in advance, and project query relation embedding $\textbf{r} \in \textbf{R}$ to $r$-aware relation representation which is used in message generating. 
Formally, $\textbf{w}_r(z,p,o_i)$ can be derived as Equation~\ref{eq_rel_project}, thus considering awareness of query relation $r$.
\begin{equation}\label{eq_rel_project}
\textbf{w}_r(z,p,o_i) = \textbf{W}_p \textbf{r}+\textbf{b}_p
\end{equation}

For $\textsc{Msg}$ function, we use vectorized approach of multiplication ~\cite{yang2014embedding} as Equation \ref{eq_msgfun}, where operator $\otimes$ is defined as element-wise multiplication between $\textbf{h}$ and $\textbf{w}$. The operation of multiplication can be interpreted as scaling $\textbf{h}_{z}^{t,l-1}$ by $\textbf{w}_r(z,p,o_i)$ in our path information modeling. 
\begin{equation}\label{eq_msgfun}
\textsc{Msg}\big(\textbf{h}_{z}^{t,l-1},\textbf{w}_r(z,p,o_i)\big) = 
\textbf{h}_{z}^{t,l-1} \otimes \textbf{w}_r(z,p,o_i)
\end{equation}

And for $\textsc{Agg}$ function, we adopt principal neighborhood aggregation (PNA) proposed in~\cite{corso2020principal}, since previous work has verified its effectiveness~\cite{zhu2022neural}.
After $w$-layers iteration of aggregation, the status of path memory $\textbf{m}_{(s,r)\rightarrow {o_i}}^{t} \in \textbf{M}_{(s,r)\rightarrow {\mathcal{E}}}^{t}$ is finally updated as $\textbf{h}_{o_i}^{t,w} \in \textbf{H}_{\mathcal{E}}^{t,w}$ w.r.t. the object candidate $o_i \in \mathcal{E}$.

As mentioned in Equation \ref{eq_overview}, $\textbf{H}_{\mathcal{E}}^{t,0}$ is initialized by memory passing strategy using last adjacent memory status $\textbf{M}_{(s,r)\rightarrow {\mathcal{E}}}^{t-1}$. Different from common GNN-based model and previous approach, for the first history subgraph (i.e. $t=0$) which has no status of previous memory, we initialize query relation embedding $\textbf{r}$ on entity $o_i$ before the first layer only if $o_i$ equals to query subject $s$, i.e. $\textbf{h}_{s}^{0,0} \leftarrow \textbf{r}$, and a zero embedding otherwise. For the case where $t>0$, memory passing strategy is involved to set the initial status $\textbf{H}_{\mathcal{E}}^{t,0}$ of PAU iteration.

\subsection{Memory Passing Strategy}\label{method_passing}

In order to capture the path information across time, we continue to refine the previously maintained path information. Considering the change of adjacent topology,
a time-aware memory passing strategy is designed to guide how much previous obtained path memory information to keep and forget.

As mentioned at the end of Section \ref{method_pau}, $\textbf{H}_{\mathcal{E}}^{0,0}$ is initialized by query relation representation at the beginning of the time ($t=0$). For $t>0$, the initial status $\textbf{H}_{\mathcal{E}}^{t,0}$ should inherit previous memory status $\textbf{M}_{(s,r)\rightarrow {\mathcal{E}}}^{t-1}$, i.e. the final layer status $\textbf{H}_{\mathcal{E}}^{t-1,w}$, which has already collected the path information until timestamp $t-1$. Therefore, we design a time-aware gated memory passing strategy, 
and the initial status $\textbf{H}_{\mathcal{E}}^{t,0}$ is determined by two parts, namely, 
the very beginning initial status $\textbf{H}_{\mathcal{E}}^{0,0}$ and 
last learned memory status $\textbf{M}_{(s,r)\rightarrow {\mathcal{E}}}^{t-1}$ at timestamp $t-1$. Formally,
\begin{equation}
\textbf{H}_{\mathcal{E}}^{t,0} = \textbf{U}_t \odot \textbf{H}_{\mathcal{E}}^{0,0}
+
(1- \textbf{U}_t) \odot \textbf{M}_{(s,r)\rightarrow {\mathcal{E}}}^{t-1},
\end{equation}

\noindent where $\odot$ denotes the Hadamard product operation. The time-aware gate matrix $\textbf{U}_t \in \mathbb{R}^{|\mathcal{E}| \times d}$ applies a non-linear transformation as:
\begin{equation}
\textbf{U}_t = \sigma(
\textbf{W}_{\textsc{Gate}}\textbf{M}_{(s,r)\rightarrow {\mathcal{E}}}^{t-1} + \textbf{b}_\textsc{Gate}
),
\end{equation}

\noindent where $\sigma(\cdot)$ denotes the sigmoid function and $\textbf{W}_{\textsc{Gate}} \in \mathbb{R}^{d \times d}$ is the parameter of time-aware gate weight matrix.

\subsection{Learning and Inference}\label{method_learning}
DaeMon models the query-aware path information across history time with the path memory. In contrast to most previous models, we focus on gathering the edge features of paths in the whole modeling process, without considering any node embedding. Thus, DaeMon has the transferability to migrate to other datasets for TKG prediction using pre-trained models, which will be presented in Section \ref{exp_result} in detail.

\noindent\textbf{Score Function.}\quad Here we show how to apply the final learned memory status $\textbf{M}_{(s,r)\rightarrow {\mathcal{E}}}^{T}$ to the TKG future reasoning.
Given query subject $s$ and query relation $r$ at timestamp $T+1$, we predict the conditional likelihood of the future object candidate $o_i \in \mathcal{E}$ using $\textbf{m}_{(s,r)\rightarrow {o_i}}^{T} \in \textbf{M}_{(s,r)\rightarrow {\mathcal{E}}}^{T}$ as:
\begin{equation}
    p(o_i|s,r)=\sigma(\mathcal{F}(\textbf{m}_{(s,r)\rightarrow {o_i}}^{T})),
\end{equation}
\noindent where $\mathcal{F(\cdot)}$ is a feed-forward neural network and $\sigma(\cdot)$ is the sigmoid function. As we have added inverse quadruple $(o,r^{-1},s,t)$ into the dataset in advance, without loss of generality, we can also predict subject $s_i \in \mathcal{E}$ given query relation $r$ and query object $o$ with the same model as:
\begin{equation}
    p(s_i|o,r^{-1})=\sigma(\mathcal{F}(\textbf{m}_{(o,r^{-1})\rightarrow {s_i}}^{T})).
\end{equation}

\noindent\textbf{Parameter Learning.}\quad Reasoning on a given query can be seen as a binary classification problem. We minimize the negative log-likelihood of positive and negative triplets as Equation \ref{eq_loss}. Negative samples are generated at each reasoning timestamp according to Partial Completeness Assumption ~\cite{galarraga2013amie}. That is we corrupt one of the entities in a positive triplet
to create a negative sample from the reasoning future triplet set.
\begin{equation}\label{eq_loss}
\mathcal{L}_{TKG} = 
-\log p(s,r,o)- \sum_{j=1}^{n} \frac{1}{n} log(1-p(\overline{s_j},r,\overline{o_j})),
\end{equation}
where $n$ is hyperparameter of negative samples number per positive sample; $(s,r,o)$ and $(\overline{s_j},r,\overline{o_j})$ are the positive sample and $j$-th negative sample, respectively.

To encourage orthogonality in the learnable parameter $\textbf{R}$ initialized at the beginning, according to ~\cite{xu2020variational}, a regularization term is also added to the objective function as Equation~\ref{eq_regloss}, where $\textbf{I}$ is the identity matrix, $\alpha$ is a hyperparameter and $\|\cdot\|$ denotes the L2-norm.
\begin{equation}\label{eq_regloss}
\mathcal{L}_{REG} = \|\textbf{R}^\top\textbf{R}-\alpha\textbf{I}\|
\end{equation}

Therefore, the final loss of DaeMon can be denoted as:
\begin{equation}
    \mathcal{L}=\mathcal{L}_{TKG}+\mathcal{L}_{REG}.
\end{equation}

\section{Experiments}

\subsection{Experimental Setup}

\begin{table}
    \centering
    \footnotesize
    \setlength\tabcolsep{2.1pt}  
    \begin{tabular}{ccccccc}
    \toprule
       Datasets  & $|\mathcal{E}|$ & $|\mathcal{R}|$ & $N_{train}$ & $N_{valid}$ & $N_{test}$ & $N_{timestamp}$\\
    \midrule
    ICEWS18 & 23,033 & 256 & 373,018 & 45,995 &  49545 & 304 \\
    GDELT & 7,691 & 240 & 1,734,399 &  238,765 & 305,241 & 2976 \\
    WIKI & 12,554 & 24 & 539,286 & 67,538 & 63,110 & 232 \\
    YAGO & 10,623 & 10 & 161,540 & 19,523 & 20,026 & 189 \\
    \bottomrule
    \end{tabular}
    \caption{Statistics of Datasets ($N_{train}$, $N_{valid}$, $N_{test}$ are the numbers of facts in training, validation, and test sets.).}
    \label{tab:data_statistics}
    \vspace{-4mm}
\end{table}

\noindent\textbf{Datasets.}\quad
Extensive experiments are conducted on four typical TKG datasets, namely, ICEWS18~\cite{jin2020Renet}, GDELT~\cite{2013GDELT}, WIKI~\cite{2018Deriving}, and YAGO~\cite{2013YAGO3}. 
ICEWS18 is from the Integrated Crisis Early Warning System~\cite{DVN/28075_2015} and GDELT~\cite{jin2019recurrent}  is from the Global Database of Events, Language, and Tone.
WIKI~\cite{2018Deriving}  and YAGO~\cite{2013YAGO3} are two knowledge base that contains facts with time information, and we use the subsets with a time granularity of years. 
The details of the datasets are provided in Table \ref{tab:data_statistics}.   
We also adopt the same strategy of dataset split as introduced in ~\cite{jin2020Renet} and split the dataset into train/valid/test by timestamps that (timestamps of the train) $<$ (timestamps of the valid) $<$ (timestamps of the test).

\begin{table*}[!htbp]
    \centering
    \scriptsize
    \setlength\tabcolsep{5.5pt}
    \begin{tabular}{c|cccc|cccc|cccc|cccc}
    \toprule
    \multirow{2.7}{*}{Model} &  \multicolumn{4}{c|}{ICEWS18} & \multicolumn{4}{c|}{GDELT} & \multicolumn{4}{c|}{WIKI} & \multicolumn{4}{c}{YAGO} \\

    \cmidrule{2-17}

    & MRR & H@1 & H@3 & H@10 & MRR & H@1 & H@3 & H@10 & MRR & H@1 & H@3 & H@10 & MRR & H@1 & H@3 & H@10 \\ 
    \midrule
    DistMult & 11.51 & 7.03 & 12.87 & 20.86 & 8.68 & 5.58 & 9.96 & 17.13 & 10.89 & 8.92 & 10.97 & 16.82 & 44.32 & 25.56 & 48.37 & 58.88 \\ 
    ComplEx & 22.94 & 15.19 & 27.05 & 42.11 & 16.96 & 11.25 & 19.52 & 32.35 & 24.47 & 19.69 & 27.28 & 34.83 & 44.38 & 25.78 & 48.20 & 59.01 \\ 
    ConvE & 24.51 & 16.23 & 29.25 & 44.51 & 16.55 & 11.02 & 18.88 & 31.60 & 14.52 & 11.44 & 16.36 & 22.36 & 42.16 & 23.27 & 46.15 & 60.76 \\ 
    RotatE & 12.78 & 4.01 & 14.89 & 31.91 & 13.45 & 6.95 & 14.09 & 25.99 & 46.10 & 41.89 & 49.65 & 51.89 & 41.28 & 22.19 & 45.33 & 58.39 \\ 
    \midrule
    TTransE & 8.31 & 1.92 & 8.56 & 21.89 & 5.50 & 0.47 & 4.94 & 15.25 & 29.27 & 21.67 & 34.43 & 42.39 & 31.19 & 18.12 & 40.91 & 51.21  \\ 
    TA-DistMult & 16.75 & 8.61 & 18.41 & 33.59 & 12.00 & 5.76 & 12.94 & 23.54 & 44.53 & 39.92 & 48.73 & 51.71 & 54.92 & 48.15 & 59.61 & 66.71  \\ 
    DE-SimplE & 19.30 & 11.53 & 21.86 & 34.80 & \underline{19.70} & 12.22 & \underline{21.39} & 33.70 & 45.43 & 42.60 & 47.71 & 49.55 & 54.91 & 51.64 & 57.30 & 60.17  \\ 
    TNTComplEx & 21.23 & 13.28 & 24.02 & 36.91 & 19.53 & 12.41 & 20.75 & 33.42 & 45.03 & 40.04 & 49.31 & 52.03 & 57.98 & 52.92 & 61.33 & 66.69 \\ 
    \midrule
    TANGO-Tucker & 28.68 & 19.35 & 32.17 & 47.04 & 19.42 & 12.34 & 20.70 & 33.16 & 50.43 & 48.52 & 51.47 & 53.58 & 57.83 & 53.05 & 60.78 & 65.85  \\ 
    TANGO-DistMult & 26.65 & 17.92 & 30.08 & 44.09 & 19.20 & 12.17 & 20.40 & 32.78 & 51.15 & 49.66 & 52.16 & 53.35 & 62.70 & 59.18 & 60.31 & 67.90  \\ 
    CyGNet & 24.93 & 15.90 & 28.28 & 42.61 & 18.48 & 11.52 & 19.57 & 31.98 & 33.89 & 29.06 & 36.10 & 41.86 & 52.07 & 45.36 & 56.12 & 63.77  \\ 
    RE-NET & 28.81 & 19.05 & 32.44 & 47.51 & 19.62 & \underline{12.42} & 21.00 & \underline{34.01} & 49.66 & 46.88 & 51.19 & 53.48 & 58.02 & 53.06 & 61.08 & 66.29  \\ 
    RE-GCN & \underline{30.58} & 21.01 & 34.34 & \underline{48.75} & 19.64 & 12.42 & 20.90 & 33.69 & \underline{77.55} & \underline{73.75} & \underline{80.38} & \underline{83.68} & 84.12 & 80.76 & 86.30 & 89.98  \\ 
    TITer & 29.98 & \underline{22.05} & 33.46 & 44.83 & 15.46 & 10.98 & 15.61 & 24.31 & 75.50 & 72.96 & 77.49 & 79.02 & \underline{87.47} & \underline{84.89} & \underline{89.96} & \underline{90.27}  \\ 
    xERTE & 29.31 & 21.03 & \underline{33.40} & 45.60 & 18.09 & 12.30 & 20.06 & 30.34 & 71.14 & 68.05 & 76.11 & 79.01 & 84.19 & 80.09 & 88.02 & 89.78 \\ 
    \midrule
    DaeMon & \textbf{31.85} & \textbf{22.67} & \textbf{35.92} & \textbf{49.80} & \textbf{20.73} & \textbf{13.65} & \textbf{22.53} & \textbf{34.23} & \textbf{82.38} & \textbf{78.26} & \textbf{86.03} & \textbf{88.01} & \textbf{91.59} & \textbf{90.03} & \textbf{93.00} & \textbf{93.34} \\ 

    \bottomrule
    \end{tabular}
    \caption{Overall performance comparison of different methods. The best results are highlighted in \textbf{bold}. And the second-best results are highlighted in \underline{underline}. (Higher values indicate better performance.)}
    \label{tab:overall_results}
    \vspace{-2mm}
\end{table*}

\vspace{1em}
\noindent\textbf{Evaluation Metrics.}\quad 
To evaluate the performance of the proposed model for TKG reasoning, we also choose the widely used task of link prediction on future timestamps. 
Mean Reciprocal Rank (MRR) and Hits@\{1, 3, 10\} are reported as the performance metrics to evaluate the proposed model's performance. 
Additionally, traditional filtered setting used in ~\cite{bordes2013translating}~\cite{jin2020Renet}~\cite{zhu2021learning} removes all valid quadruples that appear in the training, validation or test sets from the ranking list of corrupted facts, which is not appropriate for temporal reasoning tasks. 
Actually, only the facts occurring at the same time should be filtered. Thus, we use the time-aware filtered setting to calculate the results in a more reasonable way, which is consistent with the recent works \cite{2021TimeTraveler}\cite{han2021learning}.

\vspace{1em}
\noindent\textbf{Baseline Methods.}\quad 
We compare our proposed model with three kinds of baselines: 
(1)\textit{{KG Reasoning Models.}} Without considering the timestamps, DistMult\cite{yang2014embedding}, ComplEx~\cite{2016Complex}, ConvE~\cite{2017Convolutional}, and RotatE~\cite{2019RotatE} are compared. 
(2)\textit{{Interpolated TKG Reasoning models.}} 
We compare four interpolated TKG reasoning methods with proposed model, including TTransE~\cite{2018Deriving}, TA-DistMult~\cite{A2018Learning}, DE-SimplE~\cite{2020Diachronic}, and TNTComplEx~\cite{2020Tensor}. 
(3)\textit{{Extrapolated TKG Reasoning models.}}
We choose state-of-the-art extrapolated TKG reasoning methods, including TANGO-Tucker~\cite{han2021learning}, TANGO-DistMult~\cite{han2021learning}, CyGNet~\cite{zhu2021learning}, RE-NET~\cite{jin2020Renet}, RE-GCN~\cite{li2021temporal}, TITer~\cite{sun2021timetraveler}, and xERTE~\cite{han2020explainable}.

\vspace{1em}
\noindent\textbf{Implementation Details.}\quad 
For the Memory and PAU, the embedding dimension $d$ is set to 64; the number of path aggregation layers $w$ is set to 2; the activation function of aggregation is $relu$. Layer normalization and shortcut are conducted on the aggregation layers. 
For the Memory Passing, we perform grid search on the lengths of historical subgraph, and analysis it in detail presented in Figure \ref{fig:history_length}. 
We present the overview results with the lengths 25, 15, 10, 10, corresponding to the dataset ICEWS18, GDELT, WIKI, and YAGO in Table \ref{tab:overall_results}. 
For the parameter learning, negative sample number is set to 64; the hyperparameter $\alpha$ in regularization term is set to 1. Adam~\cite{kingma2014adam} is adopted for parameter learning with the learning rate of ${5e-4}$, and the max epoch of training is set to 30. 
We also conduct the migration experiments which use the pre-trained model to predict the facts in other datasets, to present the transferability of our proposed model. 
All experiments are conducted with EPYC 7742 CPU, and 8 TESLA A100 GPUs. In addition, we also have released multi-device parallel version code to accelerate the training and inference. 
Codes and datasets are all available at https://github.com/hhdo/DaeMon. 

\subsection{Experimental Results}
\label{exp_result}
The experiment results on TKG reasoning task are shown in Table \ref{tab:overall_results} in terms of MRR and Hits@\{1,3,10\}. 
It can be seen that the results convincingly verify the effectiveness and DaeMon can consistently outperform all the baseline methods on the four TKG datasets. Especially on WIKI and YAGO, DaeMon achieves the most significant improvements of 4.8\% and 4.1\% absolute in MRR compared with state-of-the-art method, respectively.
Specifically, DaeMon significantly outperforms all static models (those presented in the first block) and the temporal models for the interpolation setting (those presented in the second block) since DaeMon considers the time information of factors and models the subgraph sequence across timeline. It can thus capture the temporal pattern for the TKG reasoning tasks. 
For the temporal models under the extrapolation setting (those presented in the third block), 
there is no doubt that DaeMon also achieves better results because it can model the structure information and temporal pattern in a simultaneous manner, and DaeMon focuses on relation chain modeling which has more stable expression, considering the query-aware path information, using 3 key units to accurately do a more relevant search on the query over the timeline. In contrast to the previous model, node representation need not be considered, so for the TKG task, DaeMon can natively support model migration if the relation types are the same.

\subsection{History Length Analysis}
\begin{figure}
\centering
\subfigure[ICEWS18]{
\includegraphics[width=0.48\linewidth]{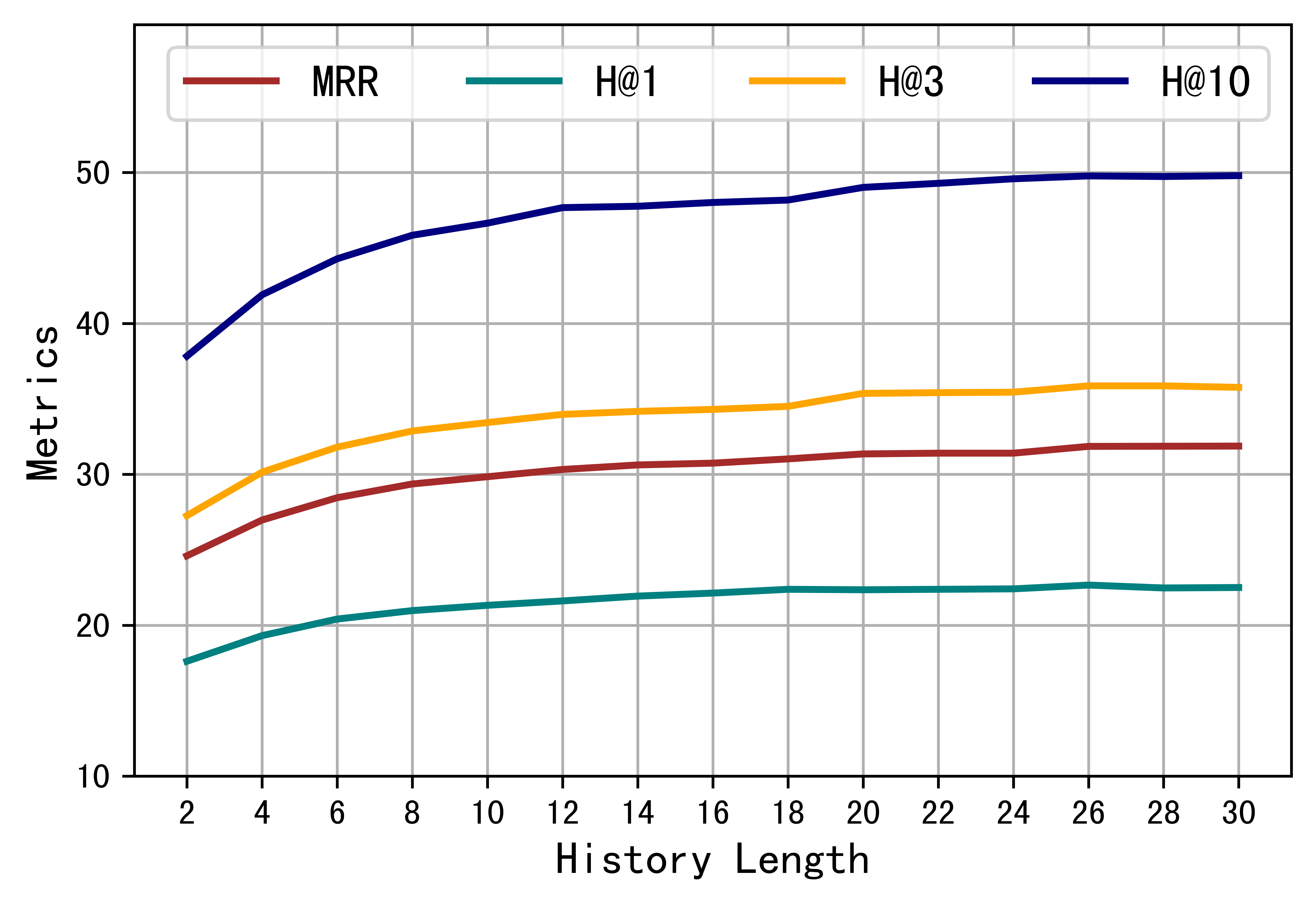}
}
\hspace{-4mm}
\vspace{-3mm}
\subfigure[GDELT]{
\includegraphics[width=0.48\linewidth]{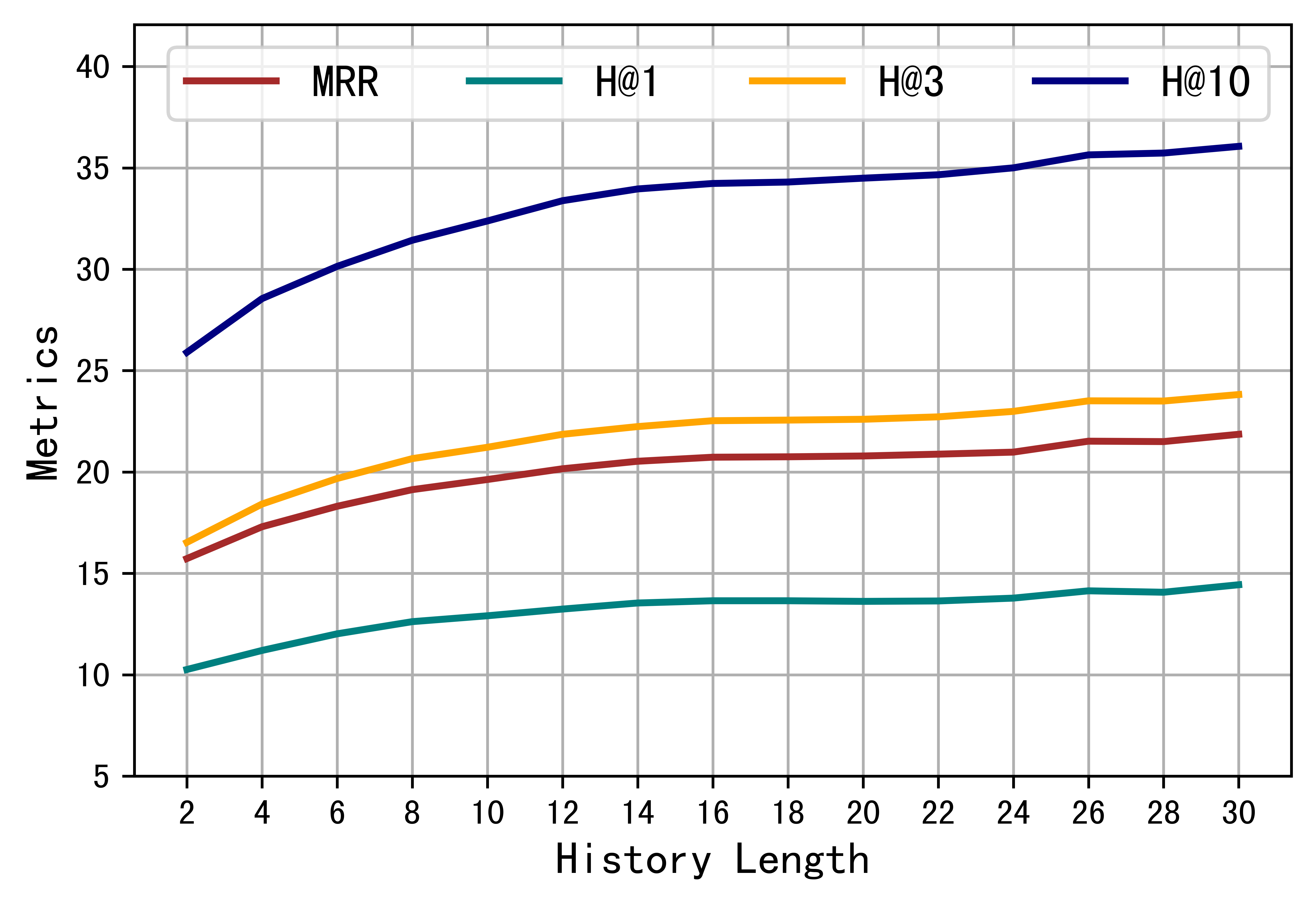}
}
\vspace{-3mm}
\subfigure[WIKI]{
\includegraphics[width=0.48\linewidth]{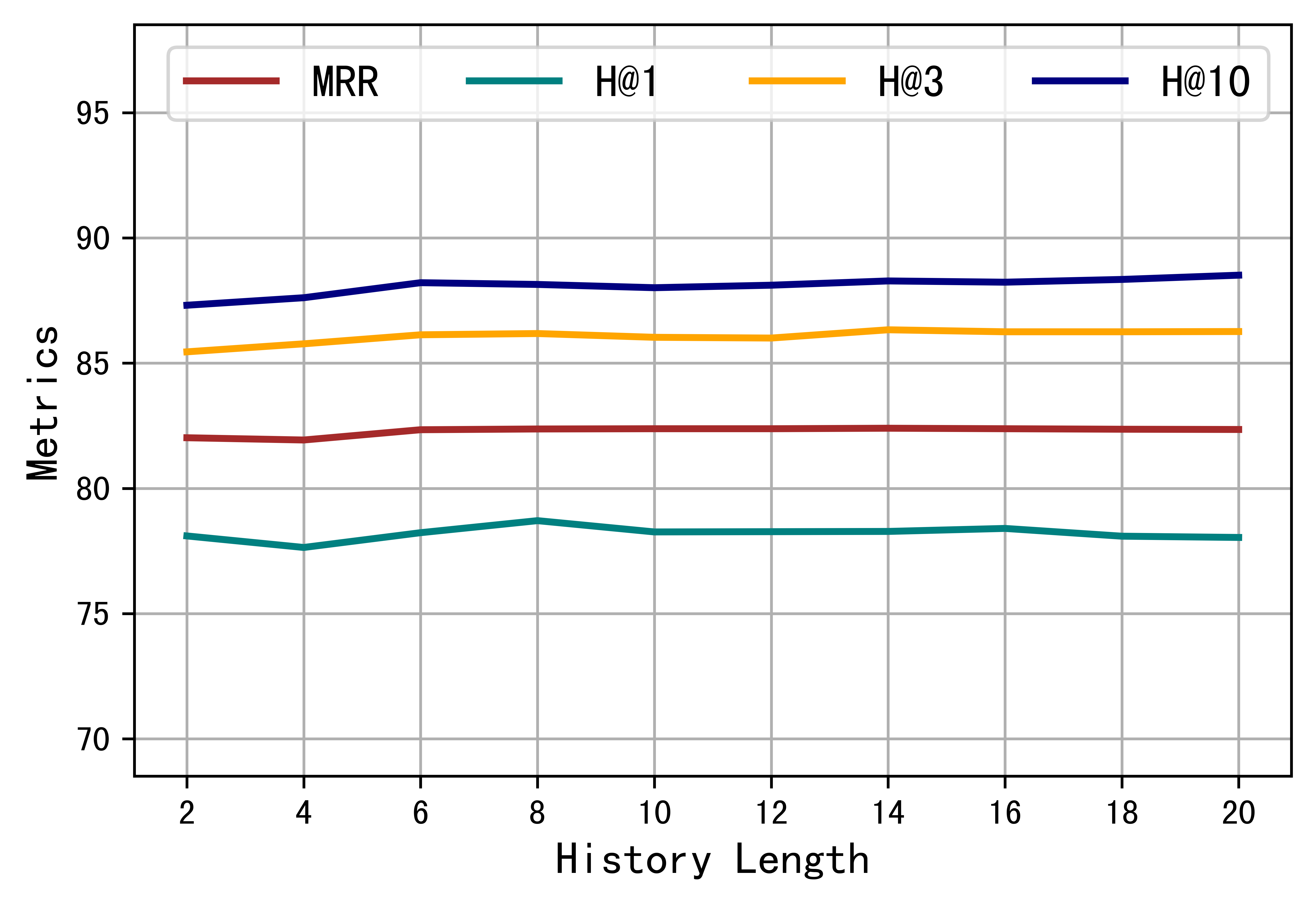}
}
\hspace{-4mm}
\subfigure[YAGO]{
\includegraphics[width=0.48\linewidth]{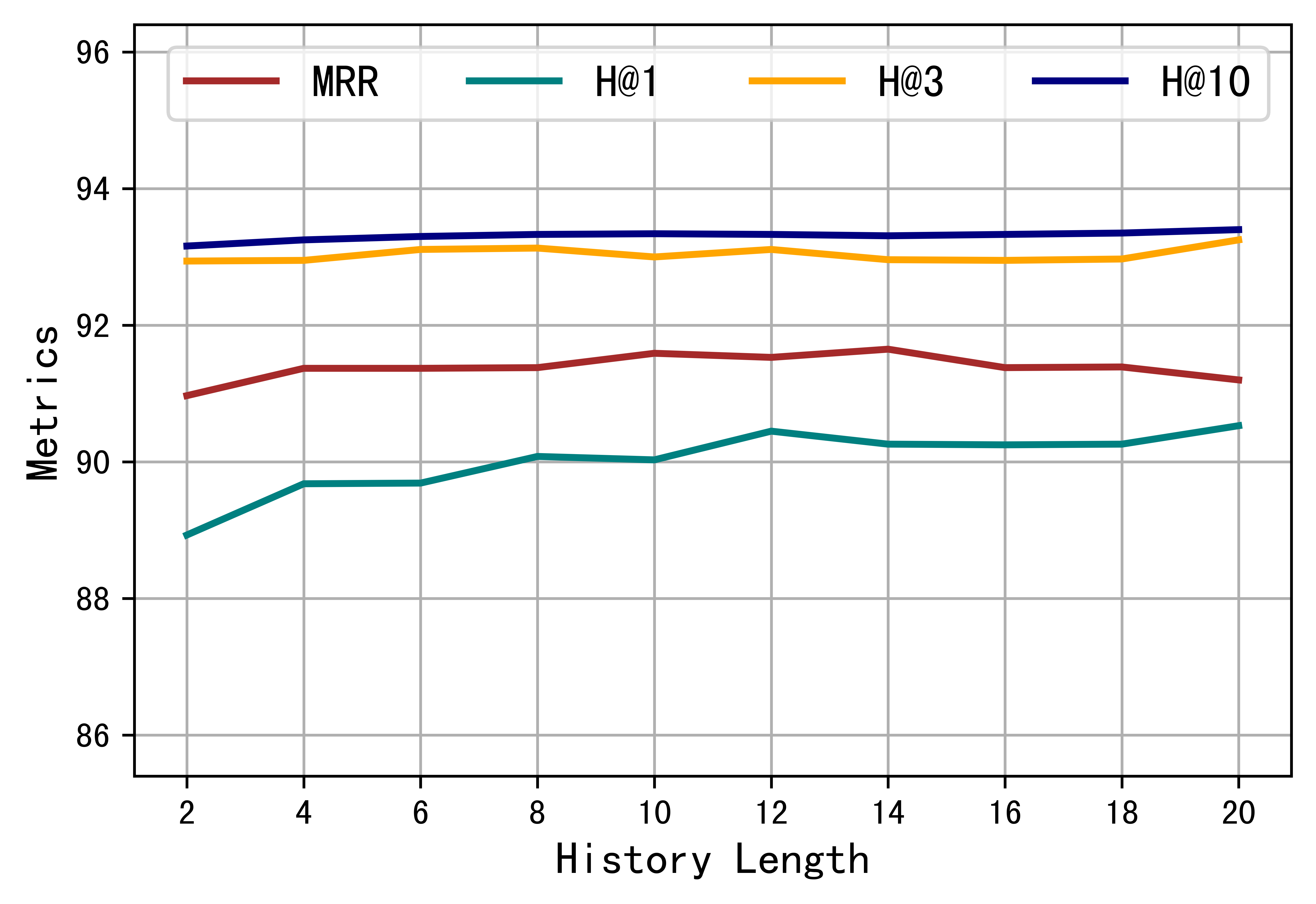}
}
\caption{Performance of Different History Length.}
\vspace{-4mm}
\label{fig:history_length}
\end{figure}

Time-across path information is updated and refined over the timeline using memory passing strategy, based on sequence of past subgraph. We adjust the values of historical sequence length to observe the performance change of DaeMon on 4 datasets. The results are shown in Figure \ref{fig:history_length}. 

In order to collect the path information across timeline, DaeMon strongly depends on the topology connectivity of the graph. Table \ref{tab:hislen} shows the density and interval of datasets.
Since WIKI and YAGO have the long interval (1 year) between times, each subgraph of them contains more information. They can perform very stably in all history length settings, as illustrated in Figure \ref{fig:history_length}(c,d). 
For the ICEWS18 and GDELT shown in Figure \ref{fig:history_length}(a,b), their interval is much shorter than the former. Thus, they need longer histories to improve the accuracy. As shown in the results, when the history length reaches to cover all entities (i.e. $length\approx |\mathcal{E}|/|\mathcal{E}_{avg}|$), MRR will no longer be significantly improved.

\begin{table}[!ht]
    \centering
    \footnotesize
    \setlength\tabcolsep{8.2pt}
    
    \begin{tabular}{c|cccc}
    \toprule
    Datasets & ICEWS18 & GDELT & WIKI & YAGO  \\ 
    \midrule
    $|\mathcal{E}|$  & 23,033 & 7,691 & 12,554 & 10,623 \\ 
    Interval & 24 hours & 15 mins & 1 year & 1 year \\    
    $|\mathcal{E}_{avg}|$  & 986.44 & 393.18 & 2817.47 & 1190.72 \\ 
    \midrule
    $|\mathcal{E}|/|\mathcal{E}_{avg}|$ & 23.35 & 19.56 & 4.46 & 8.92 \\ 
    \bottomrule
    \end{tabular}
    \caption{Density and Interval of Datasets ($|\mathcal{E}_{avg}|$ is the average number of entities involved in each timestamp.).}
    \vspace{-4mm}
    \label{tab:hislen}
\end{table}

\subsection{Ablation Study} 
We conduct ablation studies on ICEWS18 and GDELT to facilitate generalized conclusions. Particularly, we discuss the effect of each component with the following variants:  

For memory passing strategy, as shown in the first three items of Table \ref{tab:ablation_result}:
(1) we conduct mean-pooling on the previous one memory status $\textbf{M}^{t-1}_{(s,r)\rightarrow\mathcal{E}}$, i.e. the final layer status that has already modeled the path information before current processing timestamp $t$, and directly pass to the current initial status $\textbf{H}^{t,0}_\mathcal{E}$, which is denoted as \textit{MPS/PMMP}. It can be observed that \textit{MPS/PMMP} has a more significant impact on the final results, since it damages the path information that was collected before;
(2) we drop out the passing step, and initialize each initial status $\textbf{H}^{t,0}_\mathcal{E}$ as the same of $\textbf{H}^{0,0}_\mathcal{E}$, finally calculate the mean of each updated layer status $\textbf{H}^{t,w}_\mathcal{E}$ at each time, which is denoted as \textit{MPS/MMP}. It also has undesirable results, since it ignores the sequence pattern information and process path Without considering continuity.
(3) we replace the proposed gated passing strategy to \textit{Mean} operator that calculates the mean of very beginning initial status $\textbf{H}^{0,0}_\mathcal{E}$ and previous learned memory status $\textbf{M}^{t-1}_{(s,r)\rightarrow\mathcal{E}}$ as the initial layer status $\textbf{H}^{t,0}_\mathcal{E}$ at timestamp $t$, which is denoted as \textit{MPS/IPMM}. Intuitively, it belongs to a special case of the proposed strategy, and it is difficult to automatically adapt to the diversity of time series and path features.

For path aggregation unit, as shown in the fourth and fifth items of Table \ref{tab:ablation_result}, we adopt two variants of $\textsc{Msg}$ function to compare with our proposed approach.
\textit{PAU/TransE} and \textit{PAU/RotatE} replace the operator in $\textsc{Msg}$ function with vectorized version of summation and rotation operator~\cite{bordes2013translating,2019RotatE}, respectively. It can be observed that results of them are all worse than what we used in DaeMon (i.e. multiplication operator), which proves the superiority of our method.

\begin{table}[!htbp]
\centering
\scriptsize
\setlength\tabcolsep{4.2pt}

\begin{tabular}{l|cccc|cccc}
    \toprule
    \multirowcell{2.7}[0pt][c]{Model} &  \multicolumn{4}{c|}{ICEWS18} & \multicolumn{4}{c}{GDELT} \\ 
    \cmidrule{2-9}
    & MRR & H@1 & H@3 & H@10 & MRR & H@1 & H@3 & H@10 \\ 
    \midrule
    MPS/PMMP & 20.56  & 14.63  & 22.71  & 32.11  & 14.19  & 9.38  & 14.74  & 23.18  \\ 
    MPS/MMP & 25.48  & 17.82  & 28.97  & 41.08  & 16.70  & 10.43  & 17.51  & 28.58  \\ 
    MPS/IPMM & 29.85  & 20.82  & 34.03  & 47.02  & 19.98  & 12.85  & 22.01  & 33.35  \\ 
    PAU/TransE & 30.28  & 20.81  & 34.06  & 49.10  & 20.11  & 13.19  & 21.75  & 33.48  \\ 
    PAU/RotatE & 30.85  & 21.47  & 34.45  & 49.17  & 20.30  & 13.29  & 22.13  & 33.64  \\
    \midrule
    \makecell[c]{DaeMon} & 31.85  & 22.67  & 35.92  & 49.80  & 20.73  & 13.65  & 22.53  & 34.23 \\ \bottomrule
\end{tabular}
\caption{Ablation Study on ICEWS18 and GDELT.}
\vspace{-4mm}
\label{tab:ablation_result}
\end{table}

\subsection{Case Study}

In order to present the advantage of our proposed method that modeling with relation feature rather than entity (or node) embedding, we conduct a case study that shows the migration ability of DaeMon.

We first choose a target dataset `A' and another homologous dataset `B', which means `A' and `B' have the same set of relation types. 
Second, we train the DaeMon with the training data of `A' and test the performance with the testing data of `A', and we can achieve the direct result of DaeMon on target dataset `A'. 
Then, we train the DaeMon with the training data of `B' and test the performance with the testing data of `A', and we can get the migration result of DaeMon on target dataset `A' using pre-trained model derived from dataset `B'. 
Finally, we can evaluate the migration ability by the \textit{Migration Performance}, which is calculated by the percentage ratio of the migration result divided by the direct result. 

\begin{table}[!ht]
    \centering
    \footnotesize
    \setlength\tabcolsep{4.8pt}
    
    \begin{tabular}{ccccccc}

    \toprule
       Datasets  & $|\mathcal{E}|$ & $|\mathcal{R}|$ & $N_{train}$ & $N_{valid}$ & $N_{test}$ & Interval\\
    \midrule
    YAGO & 10,623 & 10 & 161,540 & 19,523 & 20,026 & 1 year \\
    YAGOs & 10,038 & 10 & 51,205 & 10,973 & 10,973 & 1 year\\
    \bottomrule
    \end{tabular}
    \caption{Comparison and Statistics of YAGO and YAGOs.}
    \label{tab:yagos_statistics}
    \vspace{-2mm}
\end{table}

More specifically, YAGO and YAGOs~\cite{han2020explainable} are homologous datasets (comparison shown in Table \ref{tab:yagos_statistics}), and there is no intersection between the entity identifications of them. Therefore we use YAGO and YAGOs to be the target dataset in turn. 
Table \ref{tab:migrate_result} shows the results of the migration ability evaluation on datasets YAGOs and YAGO. 
We can observe that all the migration performance of the DaeMon is more than 90\%. 
Even learning from a smaller dataset YAGOs and testing on a bigger dataset YAGO, the proposed model can achieve effective performance on each TKG reasoning evaluation metric. Thus, it can indicate that DaeMon can effectively capture the temporal path information and migrate the trained model to another homologous datasets.

\begin{table}[!htbp]
    \centering
    \footnotesize
    \setlength\tabcolsep{5.0pt}
    
    \begin{tabular}{c|cccc}
    \toprule
        ~ & MRR & H@1 & H@3 & H@10  \\ 
        \midrule
        YAGOs & 53.65 & 47.68 & 59.20 & 61.09  \\ 
        YAGO $\rightarrow$ YAGOs & 50.18 & 45.46 & 54.78 & 56.88  \\ \midrule
        Migration Performance & 93.53\% & 95.34\% & 92.53\% & 93.11\%  \\ \midrule
        YAGO & 91.59 & 90.03 & 93.00 & 93.34  \\ 
        YAGOs $\rightarrow$ YAGO & 88.72 & 84.73 & 92.80 & 93.14  \\ \midrule
        Migration Performance & 96.87\% & 94.11\% & 99.78\% & 99.79\% \\ \bottomrule
    \end{tabular}
    \caption{Migration Performance ($\rightarrow$ denotes the model migration.).}
    \label{tab:migrate_result}
    \vspace{-4mm}
\end{table}

\section{Conclusion}
This paper proposed DaeMon for temporal knowledge graph reasoning, which models historical information in a novel and entity-independent manner. 
Based on modeling with relation features, DaeMon adaptively captures the temporal path information between query subject and object candidates across time by utilizing historical structural and temporal characteristics while considering the query feature. 
Extensive experiments on four benchmark datasets demonstrate the effectiveness of our method on temporal knowledge graph reasoning tasks and achieve new state-of-the-art results meanwhile with natively migration ability.

\section*{Acknowledgments}
This research was supported by the Natural Science Foundation of China under Grant No. 61836013, the grants from the Strategic Priority Research Program of the Chinese Academy of Sciences XDB38030300, the Science and Technology Development Fund, Macau SAR (File no. SKL-IOTSC-2021-2023 to Pengyang Wang), the Start-up Research Grant of University of Macau (File no. SRG2021-00017-IOTSC to Pengyang Wang),  
the Informatization Plan of Chinese Academy of Sciences (CAS-WX2021SF-0101, CAS-WX2021SF-0111), 
and the Science and Technology Service Network Initiative, Chinese Academy of Sciences (No. KFJ-STS-QYZD-2021-11-001).

\bibliographystyle{named}
\bibliography{ref}

\begin{thebibliography}{}

\bibitem[\protect\citeauthoryear{Bordes \bgroup \em et al.\egroup
  }{2013}]{bordes2013translating}
Antoine Bordes, Nicolas Usunier, Alberto Garcia-Duran, Jason Weston, and Oksana
  Yakhnenko.
\newblock Translating embeddings for modeling multi-relational data.
\newblock {\em Advances in neural information processing systems}, 26, 2013.

\bibitem[\protect\citeauthoryear{Boschee \bgroup \em et al.\egroup
  }{2015}]{DVN/28075_2015}
Elizabeth Boschee, Jennifer Lautenschlager, Sean O'Brien, Steve Shellman, James
  Starz, and Michael Ward.
\newblock {ICEWS Coded Event Data}, 2015.

\bibitem[\protect\citeauthoryear{Corso \bgroup \em et al.\egroup
  }{2020}]{corso2020principal}
Gabriele Corso, Luca Cavalleri, Dominique Beaini, Pietro Li{\`o}, and Petar
  Veli{\v{c}}kovi{\'c}.
\newblock Principal neighbourhood aggregation for graph nets.
\newblock {\em Advances in Neural Information Processing Systems},
  33:13260--13271, 2020.

\bibitem[\protect\citeauthoryear{Dasgupta \bgroup \em et al.\egroup
  }{2018}]{dasgupta2018hyte}
Shib~Sankar Dasgupta, Swayambhu~Nath Ray, and Partha Talukdar.
\newblock Hyte: Hyperplane-based temporally aware knowledge graph embedding.
\newblock In {\em Proceedings of the 2018 conference on empirical methods in
  natural language processing}, pages 2001--2011, 2018.

\bibitem[\protect\citeauthoryear{Dettmers \bgroup \em et al.\egroup
  }{2017}]{2017Convolutional}
T.~Dettmers, P.~Minervini, P.~Stenetorp, and S.~Riedel.
\newblock Convolutional 2d knowledge graph embeddings.
\newblock In {\em 32nd AAAI Conference on Artificial Intelligence (AAAI-18),
  2-7 February 2018, New Orleans, LA, USA}, 2017.

\bibitem[\protect\citeauthoryear{Gal{\'a}rraga \bgroup \em et al.\egroup
  }{2013}]{galarraga2013amie}
Luis~Antonio Gal{\'a}rraga, Christina Teflioudi, Katja Hose, and Fabian
  Suchanek.
\newblock Amie: association rule mining under incomplete evidence in
  ontological knowledge bases.
\newblock In {\em Proceedings of the 22nd international conference on World
  Wide Web}, pages 413--422, 2013.

\bibitem[\protect\citeauthoryear{García-Durán \bgroup \em et al.\egroup
  }{2018}]{A2018Learning}
A~García-Durán, Sebastijan Dumani, and M.~Niepert.
\newblock Learning sequence encoders for temporal knowledge graph completion.
\newblock 2018.

\bibitem[\protect\citeauthoryear{Goel \bgroup \em et al.\egroup
  }{2020}]{2020Diachronic}
R.~Goel, S.~M. Kazemi, M.~Brubaker, and P.~Poupart.
\newblock Diachronic embedding for temporal knowledge graph completion.
\newblock pages 3988--3995, 2020.

\bibitem[\protect\citeauthoryear{Han \bgroup \em et al.\egroup
  }{2020}]{han2020explainable}
Zhen Han, Peng Chen, Yunpu Ma, and Volker Tresp.
\newblock Explainable subgraph reasoning for forecasting on temporal knowledge
  graphs.
\newblock In {\em International Conference on Learning Representations}, 2020.

\bibitem[\protect\citeauthoryear{Han \bgroup \em et al.\egroup
  }{2021}]{han2021learning}
Zhen Han, Zifeng Ding, Yunpu Ma, Yujia Gu, and Volker Tresp.
\newblock Learning neural ordinary equations for forecasting future links on
  temporal knowledge graphs.
\newblock In {\em Proceedings of the 2021 Conference on Empirical Methods in
  Natural Language Processing}, pages 8352--8364, 2021.

\bibitem[\protect\citeauthoryear{Jin \bgroup \em et al.\egroup
  }{2019}]{jin2019recurrent}
Woojeong Jin, Meng Qu, Xisen Jin, and Xiang Ren.
\newblock Recurrent event network: Autoregressive structure inference over
  temporal knowledge graphs.
\newblock {\em arXiv preprint arXiv:1904.05530}, 2019.

\bibitem[\protect\citeauthoryear{Jin \bgroup \em et al.\egroup
  }{2020}]{jin2020Renet}
Woojeong Jin, Meng Qu, Xisen Jin, and Xiang Ren.
\newblock Recurrent event network: Autoregressive structure inference over
  temporal knowledge graphs.
\newblock In {\em EMNLP}, 2020.

\bibitem[\protect\citeauthoryear{Kingma and Ba}{2014}]{kingma2014adam}
Diederik~P Kingma and Jimmy Ba.
\newblock Adam: A method for stochastic optimization.
\newblock {\em arXiv preprint arXiv:1412.6980}, 2014.

\bibitem[\protect\citeauthoryear{Lacroix \bgroup \em et al.\egroup
  }{2020}]{2020Tensor}
T.~Lacroix, G.~Obozinski, and N.~Usunier.
\newblock Tensor decompositions for temporal knowledge base completion, 2020.

\bibitem[\protect\citeauthoryear{Leblay and Chekol}{2018}]{2018Deriving}
J.~Leblay and M.~W. Chekol.
\newblock Deriving validity time in knowledge graph.
\newblock In {\em Companion of the the Web Conference}, pages 1771--1776, 2018.

\bibitem[\protect\citeauthoryear{Leetaru and Schrodt}{2013}]{2013GDELT}
Kalev Leetaru and Philip~A Schrodt.
\newblock Gdelt: Global data on events, location and tone, 1979-2012.
\newblock 2013.

\bibitem[\protect\citeauthoryear{Li \bgroup \em et al.\egroup
  }{2021}]{li2021temporal}
Zixuan Li, Xiaolong Jin, Wei Li, Saiping Guan, Jiafeng Guo, Huawei Shen,
  Yuanzhuo Wang, and Xueqi Cheng.
\newblock Temporal knowledge graph reasoning based on evolutional
  representation learning.
\newblock 2021.

\bibitem[\protect\citeauthoryear{Li \bgroup \em et al.\egroup
  }{2022}]{li-etal-2022-transher}
Yizhi Li, Wei Fan, Chao Liu, Chenghua Lin, and Jiang Qian.
\newblock {T}ran{SHER}: Translating knowledge graph embedding with
  hyper-ellipsoidal restriction.
\newblock In {\em Proceedings of the 2022 Conference on Empirical Methods in
  Natural Language Processing}, pages 8517--8528, Abu Dhabi, United Arab
  Emirates, December 2022. Association for Computational Linguistics.

\bibitem[\protect\citeauthoryear{Mahdisoltani \bgroup \em et al.\egroup
  }{2013}]{2013YAGO3}
F.~Mahdisoltani, J.~Biega, and F.~Suchanek.
\newblock Yago3: A knowledge base from multilingual wikipedias.
\newblock 2013.

\bibitem[\protect\citeauthoryear{Ning \bgroup \em et al.\egroup
  }{2021}]{Ning2021LightCAKEAL}
Zhiyuan Ning, Ziyue Qiao, Hao Dong, Yi~Du, and Yuanchun Zhou.
\newblock Lightcake: A lightweight framework for context-aware knowledge graph
  embedding.
\newblock In {\em Pacific-Asia Conference on Knowledge Discovery and Data
  Mining}, 2021.

\bibitem[\protect\citeauthoryear{Qiao \bgroup \em et al.\egroup
  }{2020}]{qiao2020context}
Ziyue Qiao, Zhiyuan Ning, Yi~Du, and Yuanchun Zhou.
\newblock Context-enhanced entity and relation embedding for knowledge graph
  completion.
\newblock {\em arXiv preprint arXiv:2012.07011}, 2020.

\bibitem[\protect\citeauthoryear{Sanchez-Gonzalez \bgroup \em et al.\egroup
  }{2020}]{sanchez2020learning}
Alvaro Sanchez-Gonzalez, Jonathan Godwin, Tobias Pfaff, Rex Ying, Jure
  Leskovec, and Peter Battaglia.
\newblock Learning to simulate complex physics with graph networks.
\newblock In {\em International Conference on Machine Learning}, pages
  8459--8468. PMLR, 2020.

\bibitem[\protect\citeauthoryear{Schlichtkrull \bgroup \em et al.\egroup
  }{2017}]{schlichtkrull2017modeling}
Michael Schlichtkrull, Thomas~N Kipf, Peter Bloem, Rianne van~den Berg, Ivan
  Titov, and Max Welling.
\newblock Modeling relational data with graph convolutional networks.
\newblock {\em arXiv preprint arXiv:1703.06103}, 2017.

\bibitem[\protect\citeauthoryear{Shang \bgroup \em et al.\egroup
  }{2019}]{shang2019end}
Chao Shang, Yun Tang, Jing Huang, Jinbo Bi, Xiaodong He, and Bowen Zhou.
\newblock End-to-end structure-aware convolutional networks for knowledge base
  completion.
\newblock In {\em Proceedings of the AAAI Conference on Artificial
  Intelligence}, volume~33, pages 3060--3067, 2019.

\bibitem[\protect\citeauthoryear{Sun \bgroup \em et al.\egroup
  }{2019}]{2019RotatE}
Z.~Sun, Z.~H. Deng, J.~Y. Nie, and J.~Tang.
\newblock Rotate: Knowledge graph embedding by relational rotation in complex
  space.
\newblock 2019.

\bibitem[\protect\citeauthoryear{Sun \bgroup \em et al.\egroup
  }{2021a}]{2021TimeTraveler}
H.~Sun, J.~Zhong, Y.~Ma, Z.~Han, and K.~He.
\newblock Timetraveler: Reinforcement learning for temporal knowledge graph
  forecasting.
\newblock 2021.

\bibitem[\protect\citeauthoryear{Sun \bgroup \em et al.\egroup
  }{2021b}]{sun2021timetraveler}
Haohai Sun, Jialun Zhong, Yunpu Ma, Zhen Han, and Kun He.
\newblock Timetraveler: Reinforcement learning for temporal knowledge graph
  forecasting.
\newblock {\em arXiv preprint arXiv:2109.04101}, 2021.

\bibitem[\protect\citeauthoryear{Trivedi \bgroup \em et al.\egroup
  }{2017}]{trivedi2017know}
Rakshit Trivedi, Hanjun Dai, Yichen Wang, and Le~Song.
\newblock Know-evolve: Deep temporal reasoning for dynamic knowledge graphs.
\newblock In {\em international conference on machine learning}, pages
  3462--3471. PMLR, 2017.

\bibitem[\protect\citeauthoryear{Trouillon \bgroup \em et al.\egroup
  }{2016}]{2016Complex}
T.~Trouillon, J.~Welbl, S.~Riedel, Ric Gaussier, and G.~Bouchard.
\newblock Complex embeddings for simple link prediction.
\newblock {\em JMLR.org}, 2016.

\bibitem[\protect\citeauthoryear{Vashishth \bgroup \em et al.\egroup
  }{2019}]{vashishth2019composition}
Shikhar Vashishth, Soumya Sanyal, Vikram Nitin, and Partha Talukdar.
\newblock Composition-based multi-relational graph convolutional networks.
\newblock {\em arXiv preprint arXiv:1911.03082}, 2019.

\bibitem[\protect\citeauthoryear{Welling and Kipf}{2016}]{welling2016semi}
Max Welling and Thomas~N Kipf.
\newblock Semi-supervised classification with graph convolutional networks.
\newblock In {\em J. International Conference on Learning Representations (ICLR
  2017)}, 2016.

\bibitem[\protect\citeauthoryear{Wu \bgroup \em et al.\egroup
  }{2020}]{wu2020temp}
Jiapeng Wu, Meng Cao, Jackie Chi~Kit Cheung, and William~L Hamilton.
\newblock Temp: Temporal message passing for temporal knowledge graph
  completion.
\newblock {\em arXiv preprint arXiv:2010.03526}, 2020.

\bibitem[\protect\citeauthoryear{Xu \bgroup \em et al.\egroup
  }{2020}]{xu2020variational}
Peng Xu, Jackie Chi~Kit Cheung, and Yanshuai Cao.
\newblock On variational learning of controllable representations for text
  without supervision.
\newblock In {\em International Conference on Machine Learning}, pages
  10534--10543. PMLR, 2020.

\bibitem[\protect\citeauthoryear{Yang \bgroup \em et al.\egroup
  }{2014}]{yang2014embedding}
Bishan Yang, Wen-tau Yih, Xiaodong He, Jianfeng Gao, and Li~Deng.
\newblock Embedding entities and relations for learning and inference in
  knowledge bases.
\newblock {\em arXiv preprint arXiv:1412.6575}, 2014.

\bibitem[\protect\citeauthoryear{Ye \bgroup \em et al.\egroup
  }{2019}]{ye2019vectorized}
Rui Ye, Xin Li, Yujie Fang, Hongyu Zang, and Mingzhong Wang.
\newblock A vectorized relational graph convolutional network for
  multi-relational network alignment.
\newblock In {\em IJCAI}, pages 4135--4141, 2019.

\bibitem[\protect\citeauthoryear{Zhu \bgroup \em et al.\egroup
  }{2021a}]{zhu2021learning}
Cunchao Zhu, Muhao Chen, Changjun Fan, Guangquan Cheng, and Yan Zhang.
\newblock Learning from history: Modeling temporal knowledge graphs with
  sequential copy-generation networks.
\newblock In {\em Proceedings of the AAAI Conference on Artificial
  Intelligence}, volume~35, pages 4732--4740, 2021.

\bibitem[\protect\citeauthoryear{Zhu \bgroup \em et al.\egroup
  }{2021b}]{zhu2021neural}
Zhaocheng Zhu, Zuobai Zhang, Louis-Pascal Xhonneux, and Jian Tang.
\newblock Neural bellman-ford networks: A general graph neural network
  framework for link prediction.
\newblock {\em Advances in Neural Information Processing Systems}, 34, 2021.

\bibitem[\protect\citeauthoryear{Zhu \bgroup \em et al.\egroup
  }{2022}]{zhu2022neural}
Zhaocheng Zhu, Mikhail Galkin, Zuobai Zhang, and Jian Tang.
\newblock Neural-symbolic models for logical queries on knowledge graphs.
\newblock {\em arXiv preprint arXiv:2205.10128}, 2022.

\end{thebibliography}

\end{document}